\documentclass[10pt,twocolumn,letterpaper]{article}

\usepackage{3dv}
\usepackage{times}
\usepackage{epsfig}
\usepackage{graphicx}
\usepackage{amsmath}
\usepackage{amssymb}
\usepackage{multirow}
\usepackage{comment}
\usepackage{enumitem}
\usepackage{makecell}

\threedvfinalcopy 

\def\threedvPaperID{34} 

\ifthreedvfinal\fi
\begin{document}

\title{Predicting Animation Skeletons for 3D Articulated Models via Volumetric Nets\vspace{-2mm}}

\author{ 
Zhan Xu$^1$
\qquad
Yang Zhou$^1$ 
\qquad
Evangelos Kalogerakis$^1$ 
\qquad
Karan Singh$^2$ \\
\qquad $^1$ University of Massachusetts Amherst 
\qquad $^2$ University of Toronto \vspace{-2mm} 
}

\maketitle

\begin{abstract}
We present a learning method for predicting animation skeletons for input  3D models of articulated characters. In contrast to previous approaches that fit pre-defined skeleton templates or predict fixed sets of joints, our method produces an animation skeleton tailored for the   structure and geometry of the input 3D model. Our  architecture is based on a stack of hourglass modules trained on a large dataset of 3D rigged characters mined from the web. It  operates on the volumetric representation of the input 3D shapes augmented with geometric shape features that provide additional cues for joint and bone locations. Our method also enables intuitive user control of the level-of-detail for the output skeleton. Our evaluation demonstrates that our approach predicts animation skeletons that are much more similar to the ones created by humans compared to several alternatives and baselines.
\end{abstract}

\section{Introduction}

Skeleton-based representations are compact representations of shapes that are particularly useful for shape analysis, recognition, modeling, and synthesis for both computer vision and graphics applications \cite{Marr78,Dickinson:2009,Tagliasacchi16}. Shape skeletons vary in definition and representation from  precise geometric concepts, such as the medial axis \cite{Blum}, to a coarse set  of joints,
possibly connected via straight segments (bones). Such jointed skeletons have been broadly used  for object recognition \cite{Song96,Felzenszwalb:2005:PSO} and shape matching \cite{Siddiqi:1999:SGS}. Another important variation  of jointed skeletons are  the ones
that capture shape pose and  mobility of underlying parts. In computer vision, these skeletons have been widely used for pose estimation \cite{Girshick:2011:ERG,Shotton2011,Wei16,Cao2017RealtimeM2,Newell2016StackedHN} and hand gesture  recognition \cite{MoonCL18,HuangZLQX18,Ren:2013:RPH}. In computer graphics, such skeletons are  used for animating articulated characters \cite{Magnenat-Thalmann:1989,Baran:2007:ARA,Shin:2001:CPI,Boulic:1996:HKB}. Artists often hand-craft animation skeletons for 3D\ models (a process known as ``rigging''), and also specify the association of the 3D model geometry with the skeleton (known as ``skinning''). As a result, the 3D model is animated when hierarchical transformations are applied to the skeletal joints.

\begin{figure}[t!]
  \centering
  \includegraphics[width=\linewidth]{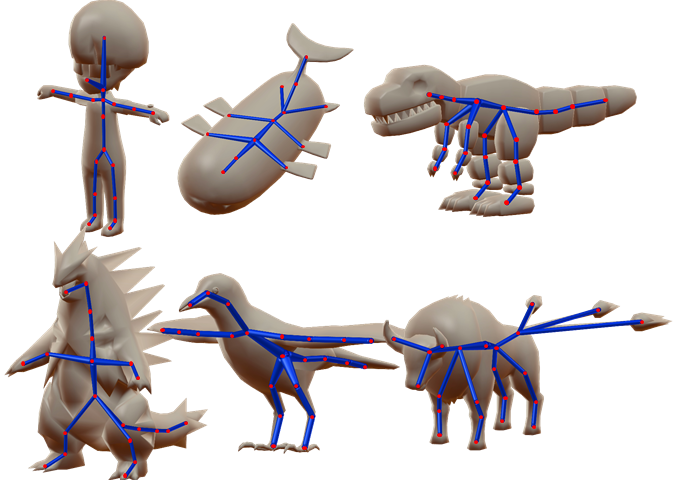}
\vspace{-4mm}    
  \caption{Examples of our predicted animation skeletons for various test 3D models. Joints are shown in red color, bones in blue.}
  \label{fig:gallery}
\vspace{-4mm}  
\end{figure}

This paper presents a deep learning approach to
predict animation skeletons of 3D\ models representing articulated characters. In contrast to existing 3D\ pose estimation methods that predict jointed skeletons for specific object classes (e.g., humans, hands) \cite{Shotton2011,MoonCL18,HuangZLQX18,PavlakosZDD17a,HaquePLAYL16,XuGZC17}, and  in contrast to existing graphics approaches that fit pre-defined skeletal templates to 3D meshes \cite{Baran:2007:ARA}, our method learns a generic model of  skeleton prediction for 3D models: it can extract plausible skeletons for a large variety of input characters, such as humanoids, quadrupeds, birds, fish, robots, and other fictional characters
(Figure \ref{fig:gallery}). Our method does not require input textual descriptions (labels) of joints, nor requires prior knowledge of the input shape category. 

\begin{figure}[t!]
  \centering
  \includegraphics[width=\linewidth]{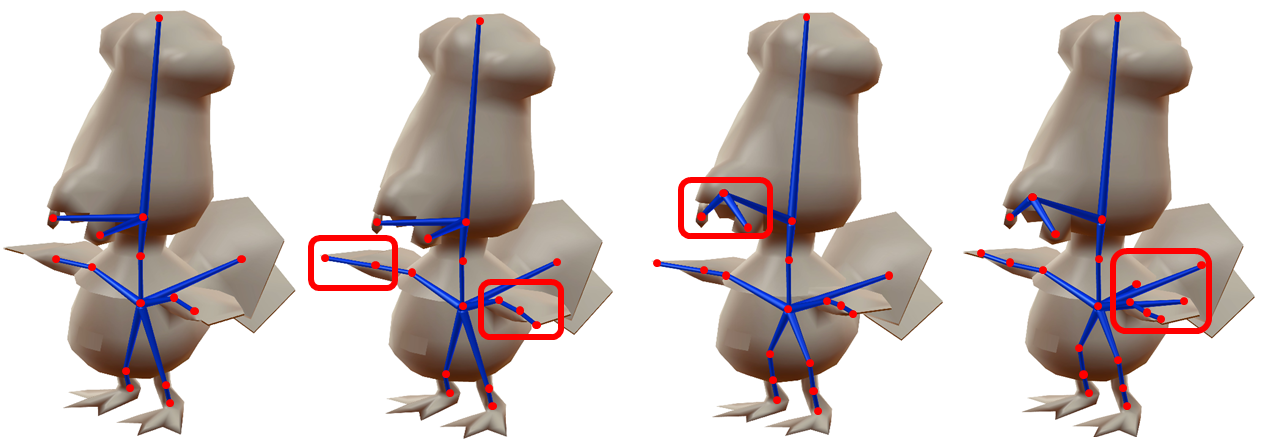}
  \vspace{-4mm}
  \caption{Effect of increasing the user parameter that controls the level-of-detail, or granularity, of our predicted skeleton. Red boxes highlight the changes in the output skeleton. }
  \label{fig:control}
  \vspace{-4mm}  
\end{figure}

There are several challenges in developing such generic approach. First, predicting jointed skeletons for a single static 3D model, without any additional information (shape class, part structure, joint labels), is under-constrained and ambiguous. To tackle this issue, we mined a large dataset of  rigged 3D characters from online  resources to train our model. 
Since the number and type of target joints and bones are  unknown for the input shape, an additional challenge for our method is to predict an appropriate skeleton tailored for the input 3D\ model such that  it captures the  mobility of its underlying articulating parts.
To form a complete animation skeleton, our method also needs to learn how to connect the predicted joints. Finally, one more challenge is to enable  user control on the granularity or level-of-detail of the output skeleton since different applications or users may require a coarser skeleton than others (Figure \ref{fig:control}). 

Our evaluation demonstrates that our method outputs skeletons that are much  closer to the ones created by human users and animators compared to several alternatives and baselines. Our contributions are the following:
\vspace{-2mm}
\begin{itemize}[leftmargin=*]
\item A\ deep  architecture that  incorporates volumetric and geometric shape features
to predict animation skeletons tailored for input
  3D models of articulated characters.
\vspace{-2mm}
\item A method to control the  level-of-detail of the output skeleton via a single, optional input parameter.
\vspace{-2mm}
\item A  dataset of rigged 3D computer character models mined from the web for training and testing learning methods for   animation skeleton prediction.   
\end{itemize}
 
\vspace{-2mm}
\section{Related Work}
Our work is most related to deep learning methods for skeleton extraction, 3D pose estimation, and character rigging. Here we provide a brief overview of these approaches.  

\vspace{-3mm}
\paragraph{Geometric skeletons.} 
Early algorithms  for skeleton extraction from 2D\ images were
based on gradients of intensity maps or distance maps to the object boundaries \cite{lindeberg_edge_1996,Siddiqi:2002,Yu04,Nedzved06,ZhangC07,Lindeberg2013}. Other traditional methods used local symmetries, Voronoi diagrams, or topological thinning as
cues for skeleton extraction \cite{Attali:1997:CSC,Amenta:1998:SRV,Liu98,Levinshtein:2013:MSP,Lee:2013:DCS,Saha:2016:SSA,tsogkas2017amat}.
As in the case of other image processing tasks, object skeleton extraction was significantly advanced by deep learning architectures.
Most deep learning approaches treat the skeleton extraction problem as a binary classification problem where the goal is to detect pixels that lie close to the medial axis of the object in the image \cite{shen2016object,shen2017deepskeleton,KeCJZY17,Zhao18,Chang18}. Alternatively, a 2D displacement or flux field can be computed from image points to geometric skeleton points \cite{Wang19}. 

Similarly to   2D traditional approaches for skeleton extraction, there has also been significant effort to extract 3D geometric skeletons, or medial surfaces (the 3D\ analog of the 2D medial axis)
 from 3D shapes. We refer the reader to \cite{Tagliasacchi16} for a recent survey. Alternatively, a\ 3D\ displacement field can  be extracted through a deep learning architecture that  maps 3D\ point sets to cross-sections of shapes, yet it cannot readily predict thin structures \cite{yin2018p2pnet}. More related to our work are methods that attempt to extract well-defined curve skeletons from 3D shapes  \cite{Tagliasacchi:2009:CSE,Cao10,Huang:2013:LMS}. However,  the resulting geometric skeletons still do not  correspond to animation skeletons i.e., their extracted segments do not  necessarily correspond to rigidly moving parts, while their extracted joints often do not lie near locations where rigid parts are connected. In addition, geometric skeletons may produce segments for non-articulating parts (i.e., parts that lack their own motion). Since our goal is to provide a  skeleton that is similar to what an animator would expect,  our training data, loss function, and  architecture  are designed to extract animation skeletons rather than geometric ones. 

\vspace{-2mm}
\paragraph{3D Pose Estimation.} Our work is also related to 3D pose estimation methods that try to recover 3D locations of joints from 2D images or directly from 3D point cloud and volumetric data 
(see also \cite{LoPresti:2016:SHA,Sarafianos2016} for related surveys). Most recent methods use deep architectures to extract joints for humans \cite{Rogez:2016:MDA,HaquePLAYL16,PavlakosZDD17a,martinez_2017_3dbaseline,Moreno2017,Zhou_2017_ICCV,Tekin2018,peng2018jointly},  hands \cite{ge2018_Point,MoonCL18,HuangZLQX18,Wan_2018_CVPR,ge2018_Point,Wan_2019_CVPR,ge2019handshapepose},
and more recently some species of animals \cite{Pereira19}. However, all these approaches aim to predict a pre-defined set of joints for a particular class of objects. In our setting, our input 3D models  drastically differ in class, structure, geometry, and number of articulating parts. Our architecture is largely inspired by the popular 2D/3D stacked hourglass networks used in pose estimation \cite{Newell2016StackedHN,MoonCL18}. However, we made several adaptations for our task, including adopting a loss function to jointly predict joints and bones and incorporating geometric features as additional cues to discover joints. Finally, since we do not assume any prior skeletal structure, we recover the underlying connectivity of the animation skeleton through a minimum spanning tree algorithm driven by our neural network. 

\vspace{-3mm}
\paragraph{Automatic Character Rigging.} A popular method for automatically extracting an animation skeleton for an input 3D model is Pinocchio \cite{Baran:2007:ARA}. The method fits a pre-defined skeleton template with a fixed set of joints to a 3D model through a combination of discrete and continuous optimization. The method can evaluate the fitting cost for different templates, and select the best one for a given model. However, hand-crafting templates to accommodate the geometric and structural variability of all possible different articulated characters is extremely hard.  Our method aims to learn a generic model of skeleton prediction without requiring any particular input templates, shape class information, or a specific set  of target joints. Our experiments demonstrate that our method predicts skeletons that are much closer to the ones created by animators compared  to Pinocchio. Recently, a neural network method was proposed to deform a 3D model based on a given input animation skeleton \cite{Liu2019}. Our  method can be used in conjunction with such skinning approaches to fully automate character rigging pipelines.

\begin{figure*}[t!]
  \centering
  \includegraphics[width=\linewidth]{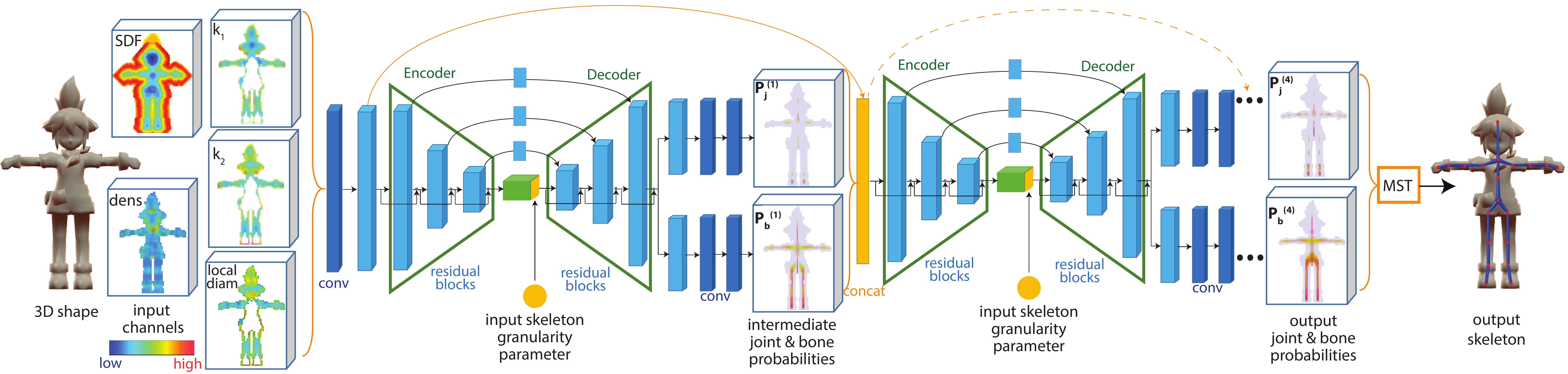}
  \caption{Pipeline of our method and deep architecture. Given an input 3D shape, we first convert it into a set of geometric representations (channels) expressed in a volumetric grid: SDF (signed distance function), LVD (local vertex density), principal surface curvatures ($k_1$, $k_2$), surface LSD (local shape diameter). We visualize cross-sections of these representations for the input shape. The input representation is processed through a stack of 3D hourglass modules. The second hourglass module is repeated two more times. Each module outputs joint and bone probabilities in the volumetric grid (visualized through cross-sections), which are progressively refined by the next module. The final joint and bone probabilities are processed through a Minimum Spanning Tree (MST)\ algorithm to extract the final skeleton. }  \label{fig:arch}
\vspace{-4mm}
\end{figure*}

\section{Overview}
Given the input geometry  of a 3D character model, our goal is to predict an animation skeleton that captures the  mobility of its underlying parts.  
 Our method has the following key components.

\begin{figure}[t!]
  \centering
  \includegraphics[width=0.95\linewidth]{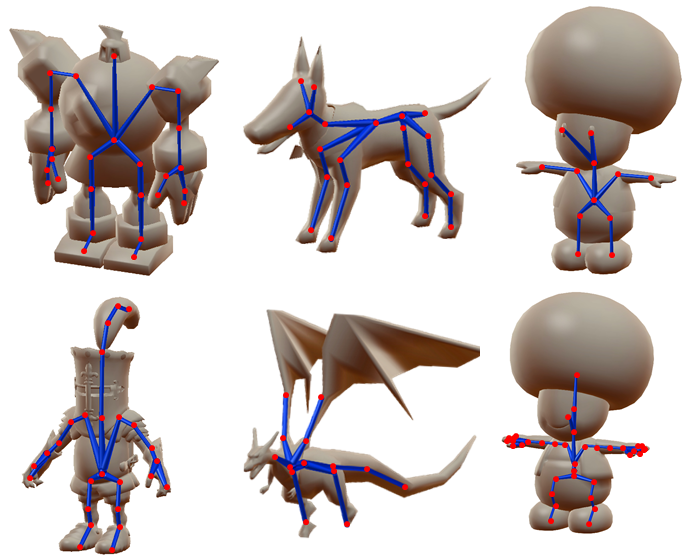}
  \vspace{-0.5mm}
  \caption{Artist-rigged 3D models from our training database.}
  \label{fig:training_db}
  \vspace{-4mm}  
\end{figure}

\vspace{-3mm}
\paragraph{Simultaneous joint and bone prediction.} In general, input characters can vary significantly in terms of structure, number and geometry of moving parts. Figure \ref{fig:training_db} shows examples of 3D models of characters rigged by artists from our collection. A single template or a fixed set of joints cannot capture such variability. Our method predicts a set of joints tailored for the input character. In addition, since the connectivity of joints is not known beforehand, and since simple connectivity heuristics based on Euclidean distance easily fail (Figure \ref{fig:mst}), our method also predicts bone segments to connect the joints. Finally, since  joint and bone predictions are not independent of each other, our method simultaneously learns to extract both through a shared stack  of encoder-decoder modules, shown in Figure \ref{fig:arch}. The stack of modules progressively refines the simultaneous prediction of bones and joints in a coarse-to-fine manner. 

\vspace{-3mm}
\paragraph{Input shape representation.} Our input 3D models are in the form of polygon mesh soups with varying topology, number of connected components, and resolution. To process them, we need to convert them into a representation that can be processed by deep networks.  Our choices of deep network and input shape representation were motivated by the fact that the target property we wish to predict, i.e., the animation skeleton,  predominantly lies in the interior of the shape. A volumetric network is well suited for this task due to its ability to make predictions away from the 3D model surface. In the context of shape reconstruction, volumetric networks \cite{maturana_iros_2015,WuSKYZTX15,riegler2017octnet} usually discretize the shape into a set of binary voxels that may lose surface detail. Following \cite{Dai17}, we instead use an implicit shape representation, namely Signed Distance Function (SDF), as input to our volumetric network. In addition, we found that additional geometric cues in the form of surface curvature, shape diameter, and mesh vertex density were also useful for our task.
 The choice of these particular geometric cues were motivated by the following observations: (a) joints are usually located near surface protrusions (e.g. knees, elbows) or concave regions (e.g., neck); principal surface curvatures are useful to characterize such areas, especially in high-resolution meshes, (b) local shape diameter \cite{Shapira:2008:CMP} changes drastically at joints where limbs are connected (e.g., hip joint); in addition, a part with constant shape diameter is usually rigged with a single bone, and (c) artist-designed 3D meshes usually include more mesh vertices near joints to promote smoother skinning and deformations; thus, local vertex density can also help to reveal joints. We found that a combination of these geometric cues with the SDF\ representation yielded the best accuracy in skeleton prediction.

\vspace{-3mm}
\paragraph{User Control.} Apart from the geometry of the input 3D model, our method also optionally takes as input a single input parameter controlling the desired granularity, or level-of-detail of the output skeleton. The reason for allowing user control  is that the choice of animations skeleton often   depends on the task. For example,  modeling crowds of characters observed from a distant camera usually does not require  rigging  small parts or extremities, such as fingers, ears and so on, since the animation of these parts would not be noticeable and would also cause additional computational overhead. In other applications, such as first-person VR environments or game environments, rigging such parts is more important. We also observed this kind of variance also in our training dataset (Figure \ref{fig:training_db}, right column):\ the skeletons of similar 3D\ models of characters differ especially near small parts  (e.g., foot, nose, and so on). Dealing with this variance is also important for learning; training with inconsistent skeletons worsens the accuracy in the joint and bone prediction. By conditioning our predictions on an input parameter capturing the desired  minimum  diameter of parts to be rigged, training converged faster and yielded skeletons closer to the ones created by artists.  We also experimented with conditioning our architecture on other parameters, such as desired minimum or average spacing between joints, yet we did not find any significant improvements.
At test time, the user can interactively change the input parameter or just use the default value, which tends to produce a moderately coarse skeleton.  

\vspace{-3mm}
\paragraph{Cross-category generalization.} Our architecture is trained on a diverse set of character categories, including humanoids, bipeds, quadrupeds, fish, toys, fictional characters, to name a few, with the goal to predict an ``as-generic-as-possible'' model, i.e., a  model that generally captures moving parts and limbs in articulated characters.

\section{Architecture}
The pipeline of our method is shown in Figure \ref{fig:arch}. It starts by converting the input 3D model into a discretized implicit surface representation augmented with geometric features.  The resulting  representation is processed through a deep network that outputs  bone and joint probabilities. The final stage  extracts the animation skeleton based on the predicted probabilities. Below we discuss the stages of our pipeline in more detail, then in the next section we discuss  training. 

\vspace{-3mm}
\paragraph{Input Shape Representation.} Our input 3D models are in the form of polygon mesh soups. Our only assumption is that they are consistently oriented.
The first stage of our pipeline is to convert them into a shape representation, which can be processed by 3D deep networks. To this purpose,  we first extract an implicit representation of the shape in the form of the Signed Distance Function (SDF) extracted through a fast marching method \cite{Sethian95afast}.  In our implementation, we use a regular $88^3$ grid.  Figure \ref{fig:arch} visualizes the SDF channel for a cross-section of an input shape.

In addition, we found that incorporating additional geometric cues increases the prediction accuracy of our method.  Specifically, we compute the two surface principal curvatures through quadratic patch fitting \cite{Hameiri2003,Goldfeather:2004:NCA} on a dense point-based sampling of the surface. We also compute the local shape diameter \cite{Shapira:2008:CMP} by casting rays opposite to the surface normals. For each volumetric cell intersecting the surface (i.e., surface voxel), we record the two principal curvatures and local shape diameter averaged across the surface points inside it. Finally, we also experimented with adding one more channel that incorporates input mesh information in the form of vertex  density. This choice is motivated by the observation that artist-designed meshes usually contain more vertices near areas that are expected to contain joints to promote smoother skinning. To incorporate this information, we perform kernel density estimation
by using a 3D Gaussian kernel centered at each mesh vertices, and record the estimated density at the center of each cell. The more vertices exist in a local 3D region, the higher the recorded density is for neighboring cells. The kernel bandwidth is set to $10$  times the average mesh edge length (estimated through grid search in a hold-out validation set). 

In total, each volumetric cell records five channels: SDF, two principal curvatures, local shape diameter, and vertex density. Thus, the resulting input shape representation $\bS$ has size $88\times 88 \times 88\times 5$.
 We note that non-surface voxels  are assigned with zero value for the two principal curvature  and local shape diameter  channels. Through volumetric convolution, the network can diffuse the surface properties in the grid, and combine them with the rest of the channels. In our supplementary material, we discuss the effects of using each of these five channels in the predicted skeletons. We also note that for different input 3D models, some channels might be more relevant than others. For example,  in the case of input meshes with near-uniform vertex density (e.g., reconstructed or re-meshed ones), the density channel is not expected to be useful. We let the learning process to weigh the input channels depending on the input accordingly

\vspace{-3mm}
\paragraph{Hourglass module.} The input shape representation is processed through a 3D hourglass network variant inspired by Huang et al. \cite{HuangZLQX18} and Moon et al. \cite{MoonCL18}. For our variant, the input shape representation is first processed through a volumetric convolutional layer and a residual block whose goal is to learn a combination of the different input features. The convolution layer has a 3D kernel of size $5\times 5\times 5$, and the residual block contains two convolutional layers with kernels $3\times 3\times 3$ and stride $1$. The output of this residual block is a new shape feature map $\bS^{(1)}$ of size $88\times 88\times 88\times 8$.
This representation is subsequent processed by an encoder module with three residual blocks that progressively  encode volumetric feature maps capturing increasingly larger and complex context in the input shape. Specifically, each of these three residual blocks consist of two volumetric convolutional layers with $3\times 3\times 3$ filters and stride $1$, and followed by another convolutional layer with stride $2$. Each stride-2 convolutional layer downsamples its input feature map by a factor of $2$. The last residual block in the encoder produces a $11\times 11\times 11\times 36$ map, which can be thought of as a compact ``code'' of the input shape.  

At this point, our architecture processes an input user parameter between $[0,1]$ corresponding to the granularity of the desired skeleton. The smaller the parameter is, the more the skeleton is extended to fine-grained, thinner parts. Figure \ref{fig:control} demonstrates the effect of varying this parameter to the skeleton. In case of no user input, a default value of $0.02$ is used (tuned through hold-out validation). The parameter is first transformed to a $11\times 11\times 11\times 4$ map, then  is concatenated with the last feature map  produced in the last residual block of the encoder resulting in a $11\times 11\times 11\times 40$ map passed to the decoder. 

The decoder is made out of 3 residual blocks that are symmetric to the encoder. Each block is followed by a transpose convolutional layer that generates a feature map of progressively increasing resolution of factor $2$. Since the feature map produced in the last residual block of the encoder encodes more global information about the shape, and may lose local details, each residual block of the decoder also accesses an earlier, corresponding feature map of the encoder after processing it through another residual block, as typically done in hourglass architectures \cite{Newell2016StackedHN}. The decoder outputs a  feature map with the same resolution as the input (size $88\times 88\times 88\times 8$). The feature map is processed by two separate branches, each consisting of a residual block and two more volumetric convolutional layers, that decrease the dimensionality of the feature maps from $8$, to $4$ and then $1$. The last feature maps from both branches are processed through a sigmoid function that outputs two probability maps individually: $\bP_j^{(1)}$ (see Figure \ref{fig:arch} for an example) represents the probability for each voxel to contain a skeletal joint, and $\bP_b^{(1)}$ represents the probability for each voxel to be on  a bone.
\vspace{-3mm}
\paragraph{Stacked hourglass network.} The predictions of joints and bones  are inter-dependent i.e., the location of joints should affect the location of bones and vice versa. To capture these inter-dependencies, we stack multiple hourglass modules to progressively refine the joint and bone predictions based on previous estimates. This stack also yielded better predictions, as we discuss in the results section. Specifically, the output maps $\{ \bP_j^{(1)},\bP_b^{(1)} \}$ are first concatenate with the shape feature presentation $\bS^{(1)}$ extracted in the first module, resulting in a $88\times 88\times 88\times 10$ representation. This is processed through a second hourglass module resulting in refined joint and bone probability maps $\{ \bP_j^{(2)},\bP_b^{(2)} \}$. These are subsequently processed by an identical third and similarly a fourth hourglass module. The last module outputs the final joint and bone probability maps 
$\{ \bP_j=\bP_j^{(4)},\bP_b=\bP_b^{(4)}\}$.  We discuss the effect of stacking  multiple modules in our results section. Details for the architecture are provided in the supplementary material. 

\begin{figure}[t!]
  \centering
  \includegraphics[width=\linewidth]{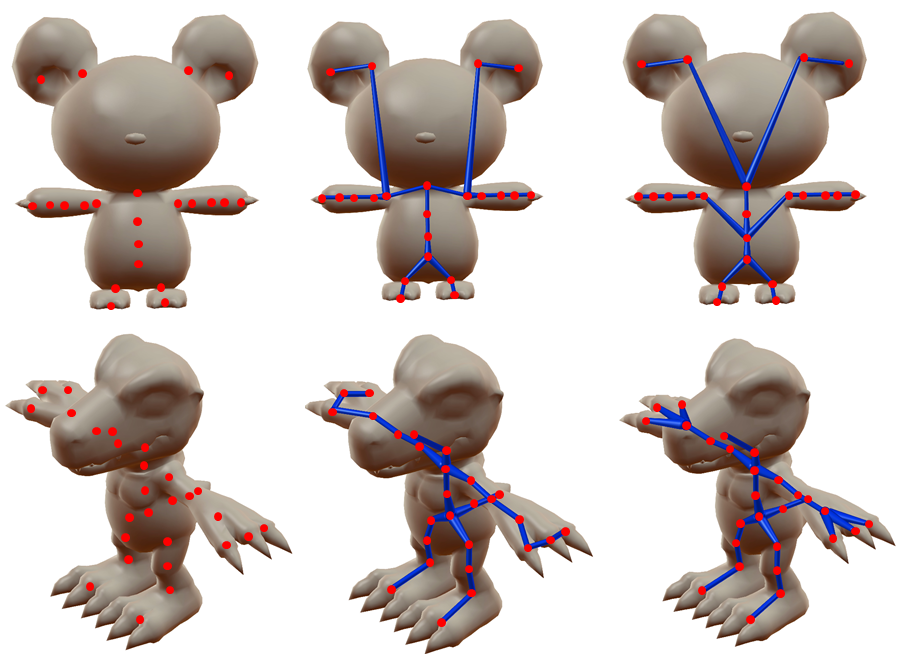}
\vspace{-4mm}  
  \caption{ \emph{Left:} Joints detected by our method. \emph{Middle:} Skeleton created with Prim's algorithm using Euclidean distance as cost. \emph{Right:} Skeleton created with Prim's algorithm using the negative log of our output bone probabilities as cost.}
\vspace{-4mm}
  \label{fig:mst}
\end{figure}

\vspace{-3mm}
\paragraph{Skeleton extraction.} The output map of joints and bones extracted from the last module of our hourglass architecture are already approximate, probabilistic indicators of the skeletal joints and bones.  As shown in the output maps in Figure \ref{fig:arch}, neighboring voxels often have correlated probabilities for joints.  To avoid multiple near-duplicate joint predictions, we apply non-maximum  suppression as a post-processing step to obtain the joints of the animation skeleton. We found that the soft non-maximum suppression procedure by Bodla et al. \cite{BodlaSCD17} is  effective at pruning non-maxima in the joint probability map. We adopt their method in our case as follows: we start with the voxel having the highest joint probability, create a skeletal joint in its position, then decay the probability of its neighborhood using a 3D isotropic Gaussian with standard deviation $\sigma=4.5$ (the deviation is tuned in a hold-out validation set). We proceed with the next voxel with the second highest probability in the updated map, create a skeletal joint, and again decay the probability of its neighborhood. The procedure stops until we cannot find any more joints with higher probability than a threshold $t=0.013$ (also tuned through hold-out validation). We also found useful to symmetrize the output probability map for symmetric characters before non-maximum suppression to ensure that the extracted joints will  be symmetric in these cases. Specifically, we first check if the input character has a global bilateral symmetry. If it does,  we reflect and average the output probability maps across the detected symmetry plane, then apply  non-maximum suppression. 

After extracting the joints, the next step is to connect them through bones. Relying on simple heuristics, such as connecting nearest neighboring joints based on Euclidean distance often fails (Figure \ref{fig:mst}a) resulting often in wrong connectivity. Instead, we use a Minimum Spanning Tree (MST) algorithm that minimizes a cost function over edges between extracted joints representing candidate skeleton bones. The MST\ algorithm also guarantees that the output animation skeleton is a tree, as typically required in graphics animation pipelines to ensure uniqueness of the hierarchical transformations applied to the bones. The cost function is driven by the bone probability map extracted by our network. If an edge between two skeletal joints intersects voxels with low bone probability, then the edge has a high cost of being a bone. If the edge instead passes through  a high bone probability region, then the cost is low. Mathematically, given an edge between joints $i$ and $j$, the 3D line segment $l_{i,j}$ connecting them, and the predicted bone probability $\bP_{b}(v)$ for each voxel that is intersected by this line segment ($v \in l_{ij}$), the cost for this edge is defined as follows:
\begin{equation*}
    w_{i,j} = -\sum_{v \in l_{i,j}} \log\bP_b(v) 
\label{path_cost}
\end{equation*}
We note that we use sum instead of the average of the bone voxel probabilities for the edge costs in the above formula. In this manner, if there are two edges that are both crossing the same high-probability bone regions, the shorter edge will be preferred since nearby joints are more likely to be connected. To prevent bones from going outside the shape, we also set the edge cost of voxels that lie outside the shape to a large number ($10^{5}$ in our implementation).

After forming a graph connecting all-pairs of joints, we  use the Prim's algorithm \cite{Prim1957} to extract the tree with a minimum total cost over its edges.
The root joint
is selected
to be the one closest to the shape centroid.
The extracted edges represent the bones of 
our animation skeleton.

\section{Training}
\begin{figure}[t!]
  \centering
  \includegraphics[width=\linewidth]{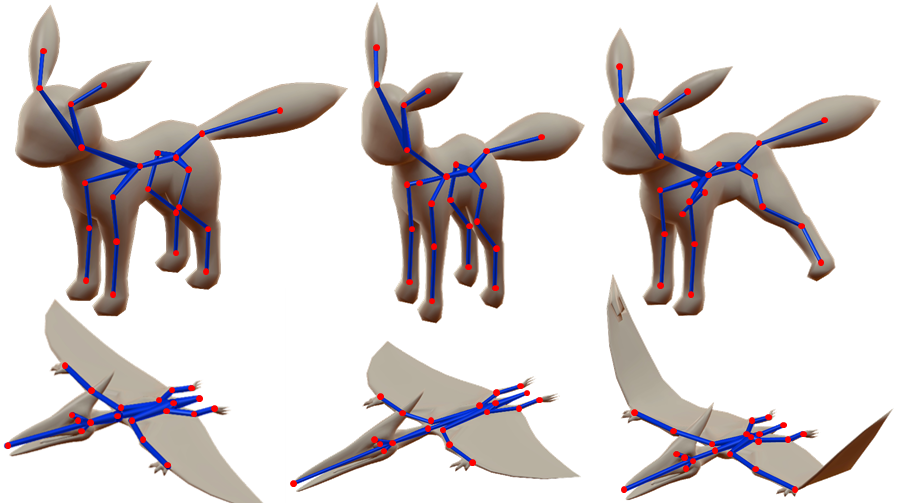}
\vspace{-4mm}  
  \caption{Examples of augmentation of our training dataset. \emph{Left:} Training 3D  models. \emph{Middle:} Generated variants through anisotropic scaling.  \emph{Right:}  Generated variants with different pose. }
\vspace{-4mm}  
  \label{fig:augmentation}
\end{figure}

Our training procedure aims to optimize the parameters of our  architecture such that the resulting probability maps agree as much as possible with the training joints and bone locations. Below we discuss the procedure to train our architecture and the details of our dataset. 
\vspace{-3mm}
\paragraph{Dataset.} We first collected a dataset of $3277$ rigged characters from an online repository, called Models Resource \cite{ModelsResource}. The models spanned various categories, including humanoids, quadrupeds, birds, fish, robots, toys, and other fictional characters. The vast majority of models are in a symmetric, neutral pose.  We excluded any duplicate models by examining both the Chamfer distance (average point-to-nearest-point distance) between the surface of the models, and also intersection over union based on their volumetric representations. The check also included varying the pose of each model. Most of the models were consistently oriented in terms of front-facing and upright axis. The rest were corrected manually (about $30\%$ of the models). The models were centered such that they are  grounded on the x-z plane, and their centroid projection on x-z plane was at $(0,0)$. They were also scaled such that  the length of their longest dimension is 1.
 
 \vspace{-3mm}
\paragraph{Splits.} After re-orientation, re-scaling and  de-duplication, our dataset contained $3193$ models. The average number of joints per character in our dataset was $26.4$. The supplementary material provides more statistics and a histogram over the number of joints across the models of our dataset. We split our dataset into $80\%$ for training ($2,554$ models), $10\%$ for hold-out validation ($319$ models), and $10\%$ for testing.  
\vspace{-3mm}
\paragraph{Augmentation.} The training split was relatively small, thus, we considered two useful types of augmentations: (a) scaling the model anisotropically along each of the 3 axes using a random scaling amount between $0.5$ and $1.5$, (b) we apply a random rotation between $30$ and $50$ degrees to joints. If two joints  found to be symmetric after checking for bilateral symmetry (e.g. hips), we apply the same random rotation to both. These augmentations promoted invariance to pose variations, and invariance to scaling. We rejected augmentations that resulted in severe, geometric self-penetrations. Examples are shown in Figure \ref{fig:augmentation}. In total, we generated up to $5$ variations of each model in our training split, resulting totally in $15,526$ training models.

\vspace{-3mm}
\paragraph{Training objective.} The training 3D models are voxelized in a $88^3$ volumetric grid, and the   input feature channels (signed distance function, principal curvatures, local shape diameter, and mesh density) are extracted for them. To train our network, we also need a value for the input granularity control parameter, which captures the minimum shape diameter of part to be rigged per training shape. We set this automatically based on the input training shape geometry and skeleton: for each model, we find the surface points nearest to its training bones and compute their local shape diameter. We set this parameter equal to the fifth percentile of the local shape diameter across these points. We used the fifth percentile instead of the minimum for robustness reasons.

Then for each training model $m$, we generate a target map for joints $\hat\bP_{v,m}$
 and  bones $\hat\bP_{b,m}$ based on their  animation skeleton. Specifically, at each joint position, we create a small 3D isotropic Gaussian heatmap with unit variance. The target joint map is created by aggregating the individual heatmaps and discretizing them into the same volumetric grid. If a voxel is affected by more than one heatmaps, we use the max value over them. This strategy of diffusing the target joint maps (instead of setting individual voxels to ones when they contain joints) led to faster convergence and more accurate skeletons at test time. Another reason for doing this is that  the skeletons were manually created by modelers, and as a result, the joint positions are not expected to be perfect. The same strategy is followed for generating the target bone map: we create a 3D isotropic Gaussian heatmap at dense samples over bones, then aggregate them and discretize them into the volumetric grid. 

The cross-entropy can be used to measure the difference  between our predicted probability maps  $\bP_{b,m}, \bP_{j,m}$ and the target heat maps $\hat\bP_{b,m}, \hat\bP_{b,m}$ for each voxel $v$. The cross entropy for joints is defined as follows:
 \begin{equation*}
 L_j[v] =  \hat\bP_j(v)\log(\bP_j(v)-(1-\hat\bP_j(v))\log(1-\bP_j(v)) 
 \end{equation*}
 The cross entropy for bones is similarly defined as follows:
 \begin{equation*}
 L_b[v] =  \hat\bP_b(v)\log(\bP_b(v)-(1-\hat\bP_b(v))\log(1-\bP_b(v)) 
 \end{equation*}
 We note that in contrast to the binary cross-entropy used in classification tasks where the target variables are binary,  in our case these are soft due to the target map diffusion. 

A large number of volumetric cells lie in the exterior of the 3D model and do not contain any joints and bones. These cells do not carry useful supervisory signal, and  can dominate the loss if we simply sum up the cross-entropy across all voxels. To prevent this, we use a masked loss. Specifically,  for each training shape $s$, we compute the mask  $\bM_s$, which is set to $1$ for voxels that are on the surface or the interior of the model, and $0$ otherwise. The final loss used to train our model is the following:

\begin{equation}
L = \sum\limits_s
\frac{1}{N_s} \sum\limits_{v} M_s[v] (L_j[v] + L_b[v]) 
\end{equation}
where $N_s = \sum_{v} M_s[v]$.  The loss is applied to the output of all  hourglass modules of our stack architecture. We also note that applying different weights for the joint and bone losses did not improve the performance.

\vspace{-3mm}
\paragraph{Optimization.} We use the Adam optimizer to minimize the  loss. Our implementation is done on PyTorch. Hyper-parameters, including the Gaussian kernel bandwith for the density channel, the parameters of the non-maximum suppresion, and variance for diffusion of target maps are set through grid search in the hold-out validation set to minimize the same loss. In addition, we choose a default value for the input user parameter also through grid search in our hold-out validation set.

\section{Results}

\begin{figure*}[t!]
  \centering
  \includegraphics[width=\linewidth]{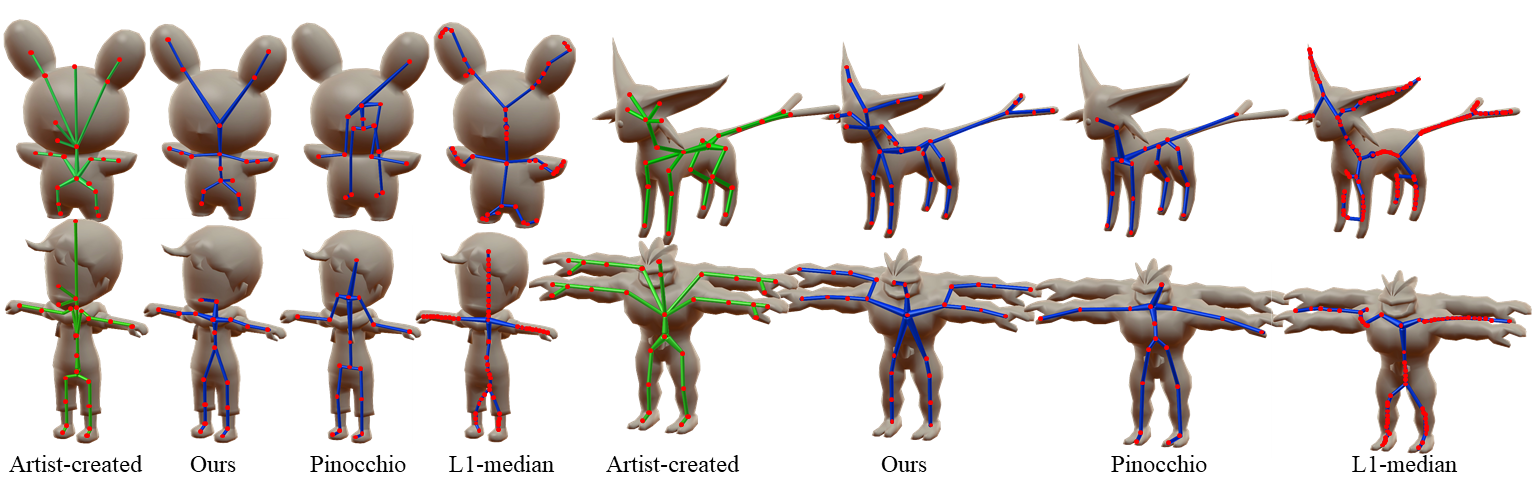}
  \caption{Comparisons of different methods for representative test characters. In each group, the green one indicates the artist-created (reference) skeleton, and the blue ones, from left to right, are our prediction, result from Pinocchio, result from the L1-median method.}
  \label{fig:comparisons}
\vspace{-2mm}
\end{figure*}

We evaluated our method and alternatives  quantitatively and qualitatively on the test split of our dataset. Below, we discuss results and comparisons.

\vspace{-1mm}
\paragraph{Quantitative evaluation measures.} The goal of our quantitative evaluation is to numerically measure the similarity of the predicted skeletons to the ones created by designers (denoted as ``reference'' skeletons in the following paragraphs).
We rely on several measures to quantify this similarity. The first evaluation measure is the Chamfer distance between joints (\emph{CD-joint}). Specifically, given a test shape, we measure the Euclidean distance from each predicted joint to the nearest joint in its reference skeleton, then compute the average over the predicted joints. The Euclidean distance is divided by the length of the longest axis in the input shape. We also compute the Chamfer distance the other way around i.e., we compute the distance from each joint in the reference skeleton to the nearest predicted joint. We then take the average of these two distance measures resulting in a symmetrized Chamfer distance. We report this distance measure averaged over our test shapes.  The higher the value is for \emph{CD-joint}, the more erroneous the placement  of the predicted joints is. 
To further characterize the misplacement, we use a second measure, which is the Chamfer distance between joints and bones (\emph{CD-joint}2bone). Given a shape, we measure the Euclidean distance from each predicted joint to the nearest bone point on the artist-created skeleton, then compute the average.  As in the case the previous measure, also we symmetrize this measure by evaluating the distance from the reference skeleton joints to the predicted bones. If \emph{CD-joint2bone} is much lower than \emph{CD-joint}, it indicates that the predicted and reference skeletons overlap, yet the joints are misplaced along the direction of the bones. Ideally, both \emph{CD-joint2bone} and \emph{CD-joint2bone} should be low.

Another evaluation measure we use is the matching rate of the predicted joints \emph{(MR-pred)}: this is defined
as the percentage of predicted joints whose distance to their nearest reference ones is lower than a prescribed tolerance. In other words, if a predicted joint is located closer  to a reference joint than this tolerance, it counts as a correct prediction. Similarly, we also define the matching rate of the reference joints \emph{(MR-ref)}. This is  the percentage of reference joints whose distance to  their nearest predicted joints is lower than the tolerance. The tolerance is normalized by the local shape diameter evaluated at the nearest reference joint. This  is computed by casting rays perpendicular to the bone starting at this joint and computing the average distance between intersection points at the surface along opposite ray directions.  The reason for using this normalization is that at increasingly thinner parts, joint misplacement becomes more pronounced e.g., a predicted joint may have low absolute distance to the nearest reference joint, but is located outside the shape.  


\vspace{-3mm}
\paragraph{Comparisons.} Our method was evaluated against the following prior works. First, we compare with \emph{Pinocchio} \cite{Baran:2007:ARA}, which fits a template skeleton selected for each input model. The template is automatically selected among a set of predefined templates (humanoid, short quadruped, tall quadruped, and centaur) by evaluating the fitting cost for each of them, and choosing the one with the minimum one. Second, we compare with the \emph{L1-medial} skeleton that computes a geometric skeleton representation for an input point set, representing its localized center. To compute the L1 medial skeleton, we uniformly sample the surface of each input shape with $1000$ points. The method also creates a set of joints through morphological operations. We tune the parameters of the method in our hold-out validation set through grid search. 

\begin{table}[t!]
\centering
\caption{Evaluation for all competing methods} 
\centering 
\begin{tabular}{|@{}c@{}| @{}c@{}| @{}c@{}| @{}c@{}| @{}c@{}| } 
\hline\hline 
Method  & \,CD-joint\, & \,CD-joint2bone\,  & \,MR-pred\, & \,MR-ref\, \\ [0.5ex] 
\hline 
Pinocchio & 7.4\% & 5.8\% & 55.8\% & 45.9\% \\ 
L1-median & 5.7\% & 4.4\% & 47.9\% & 63.2\% \\
Ours & \textbf{4.6\%} & \textbf{3.2\%} & \textbf{62.1\%} & \textbf{68.3\%} \\  
\hline 
\end{tabular}
\label{table:compare_tab} 
\vspace{-3mm}
\end{table}
Table \ref{table:compare_tab} reports our evaluation measures for all the competing methods. Our method achieves the lowest Chamfer distances (\emph{CD-joint}, \emph{CD-joint2bone}). Our \emph{CD-joint2bone} measure is also lower than \emph{CD-joint} indicating that our predicted skeletons tend to overlap more with the reference ones, and errors are mostly due to misplacing joints along the bones of the reference skeleton. In addition, for a conservative tolerance of $0.5$ of the local shape diameter, our method achieves the highest marching rates.  All evaluation measures indicate that our method produces  more accurate skeletons with respect to  the reference ones.
Figure \ref{fig:comparisons} shows the reference skeletons and  predicted ones from different methods for some characteristic test shapes. We observe that our methods tends to output skeletons whose joints and bones are closer to the artist-created ones.

\paragraph{Ablation study.} In our supplementary material (see appendix), we present evaluation of alternative choices for our method. All  the variants are trained in the same split, and tuned
in the same hold-out validation set in the same manner as our original method.  We examined the effect of  different numbers of hourglass modules in our  architecture. We observed that the performance saturates when we reach $4$ hourglass modules. We also evaluated the  effect of geometric features
and the input granularity control parameter.
We found that all are useful to increase the performance. Finally, we  examined the effect of predicting only joints and connecting them based on Euclidean distance as cost for Prim's algorithm. We observed that the performance  degrades without driving it through our bone predictions.
We refer the reader to the supplementary material for our ablation study.

\section{Conclusion}
We presented a method for learning animation skeletons for 3D computer characters. To the best of our knowledge, our method represents a first step towards learning a generic, cross-category model for producing animation skeletons of 3D models.

There are still several limitations. First, the method is based on a volumetric networks with limited resolution, which can result in missing joints for small parts, such as fingers, or misplacing other joints, such as knees and elbows. In the future, it would be interesting to investigate other networks, such as octree-based ones \cite{wang2017ocnn,riegler2017octnet,Wang:2018:AOP}, graph or mesh-based ones \cite{masci2015geodesic,Boscaini2016,Monti2017,bronstein2017geometric,Liu2019}.
The animation skeleton is produced through a post-processing stage in our method, which might result in undesirable joint connectivity (e.g., see shoulder joints for the four-armed alien of Figure \ref{fig:comparisons}). An end-to-end method would be more desirable.  
 It would also be interesting to investigate learning methods that jointly estimate skinning weights  \cite{Liu2019} and animation  skeletons.

\vspace{1mm} 
\textbf{Acknowledgements.} This research is funded by NSF (CHS-161733). Our experiments were performed in the UMass GPU cluster obtained under the Collaborative Fund managed by the Massachusetts Technology Collaborative.

{\small
\bibliographystyle{ieee}
\bibliography{egbib}
}
\newpage
\section*{Appendix: Supplementary Material}
\vspace{+2mm}
\begin{figure}[b!]
  \centering
  \includegraphics[width=0.8\linewidth]{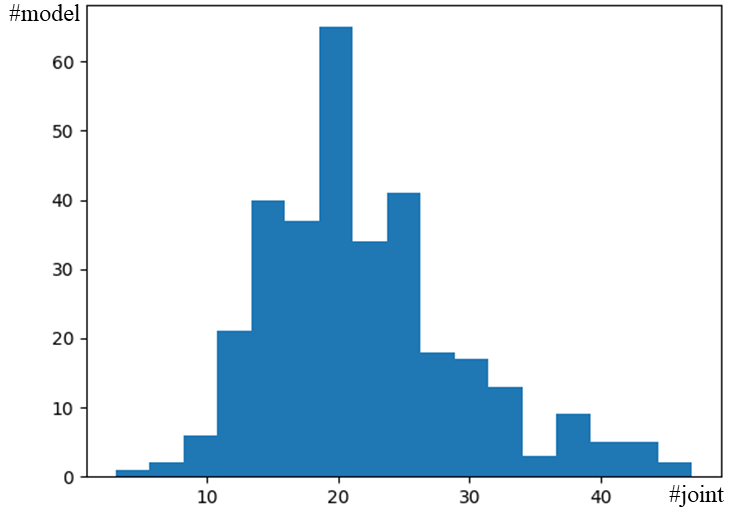}
  \caption{Histogram over the number of joints across the models of our dataset}
  \label{fig:hist}
\end{figure}

\begin{table}[b!]
\caption{Evaluation of varying number of  hourglass modules} 
\centering 
\begin{tabular}{|@{}c@{}| @{}c@{}| @{}c@{}| @{}c@{}| @{}c@{}|} 
\hline\hline 
(\#) Modules \,  & \,CD-joint\, & \,CD-joint2bone\, & \,MR-pred\, & \,MR-ref\, \\
\hline 
1 & 5.2\% & 3.4\% & 55.7\% & 60.9\% \\ 
2 & 4.9\% & 3.3\% & 60.0\% & 65.5\% \\
3 & 4.7\% & 3.3\% & 61.4\% & 67.0\% \\
4 & \textbf{4.6\%} & \textbf{3.2\%} & \textbf{62.1\%} & \textbf{68.3\%} \\ 
\hline 
\end{tabular}
\label{table:num_hg} 
\end{table}


\paragraph{Ablation Study.}

Here we present evaluation of alternative choices for our method. All  the variants are trained in the same split and tuned in the same hold-out validation set in the same manner as our original method. Table \ref{table:num_hg} reports the same evaluation measures described in Section 6 of our paper for   different number of hourglass modules in our  architecture. We observed that the performance saturates when we reach $4$ hourglass modules. 

We also evaluated the geometric features used as input to our architecture. Table \ref{table:input_feature} reports the evaluation measures when using the Signed Distance Function only (SDF), the SDF\ plus each of the other geometric features, and altogether. We can see that each geometric feature individually improves the performance, and integrating all of them achieves the best result.

Table \ref{table:condition} reports the performance when
(a) we remove the granularity control parameter from the architecture, (b) use Euclidean distances as cost for Prim's algorithm instead of the predicted log probabilities for bones. Both degraded variants drop the performance especially in terms of precision and recall. 


\paragraph{Dataset Statistics.}
Our de-duplicated dataset contained $3193$ rigged characters  from Models Resource. The average joint number per character is $26.4$.  Figure \ref{fig:hist} shows a histogram over the number of joints across the models of our dataset.


\paragraph{Architecture details.}

Table \ref{table:layers} lists each layer used in our architecture along with the size of its output map.

\begin{table}[t!]
\vspace{-2mm}
\centering 
\caption{Evaluation of different input feature combinations} 
\begin{tabular}{|@{}c@{}| @{}c@{}| @{}c@{}| @{}c@{}| @{}c@{}| } 
\hline\hline 
Input features\, & \,CD-joint\, & \,CD-joint2bone\, & \,MR-pred\, & \,MR-ref \, \\ [0.5ex] 
\hline 
SDF only & 5.2\% & 3.5\% & 60.6\% & 56.0\% \\ 
SDF+diam. & 4.9\% & 3.3\% & 53.5\% & 61.8\% \\
SDF+curv. & 4.7\% & \textbf{3.2\%} & 51.2\% & 66.4\% \\
SDF+density & 4.7\% & \textbf{3.2\%} & 57.5\% & 63.2\% \\

\hline\hline 

all features & \textbf{4.6\%} & \textbf{3.2\%} & \textbf{62.1\%} & \textbf{68.3\%} \\ 
\hline 
\end{tabular}
\label{table:input_feature} 
\end{table}

\begin{table}[t!]
\vskip -2mm
\caption{Evaluation of skipping the granularity control parameter (no control) and  using Euclidean distances instead of log bone probabilities for Prim's algorithm (no bone prob)} 
\centering 
\begin{tabular}{|@{}c@{}| @{}c@{}| @{}c@{}| @{}c@{}| @{}c@{}|} 
\hline\hline 
 ~ variant \,  & \,CD-joint\, & \,CD-joint2bone\, & \,MR-pred\, & \,MR-pref\, \\
\hline 
no  control & 4.6\% & 3.2\% & 54.5\% & 67.9\% \\
no bone prob. & 4.6\% & 3.2\% & 57.8\% & 67.0\% \\
full method  
& \textbf{4.6\%} & \textbf{3.2\%} & \textbf{62.1\%} & \textbf{68.3\%} \\
\hline 
\end{tabular}
\label{table:condition} 
\vspace{-2mm}
\end{table}

\begin{table}[h!]
\centering 
\caption{\textbf{Architecture details.} ResBlock: The residual block is made of two volumetric convolutional layers with filters 3$\times$3$\times$3. Both produce the same number of feature maps. When the number of input/output feature maps differ, the skip path within any residual block contains an additional volumetric convolutional layer with 3$\times$3$\times$3 filters. Dropout: dropout layer with 0.2 probability. }
\footnotesize
\begin{tabular}{|@{}c@{}|@{}c@{}|@{}c@{}|}
\hline
\hline

    & Layers                                            & Output   \\
\hline
\hline
    & Input volume                                      &88$\times$88$\times$88$\times$5  \\\cline{2-3}
    & ReLU(BN(Conv(5x5x5, 5$\rightarrow$8)))            &88$\times$88$\times$88$\times$8  \\\cline{2-3}
    & ResBlock                                          &88$\times$88$\times$88$\times$8  \\
\hline
\hline
\multirow{6}{*}{Encoder}  & ReLU(BN(Conv(2x2x2, stride=2))) 
                          &\makecell{44$\times$44$\times$44$\times$8 \\for 1st module, \\44$\times$44$\times$44$\times$10 \\for the rest}  \\\cline{2-3}
                          & ResBlock                    &44$\times$44$\times$44$\times$16 \\\cline{2-3}
                          & ReLU(BN(Conv(2x2x2, stride=2)))                 &22$\times$22$\times$22$\times$16 \\\cline{2-3}
                          & ResBlock                    &22$\times$22$\times$22$\times$24 \\\cline{2-3}
                          & ReLU(BN(Conv(2x2x2, stride=2)))                 &11$\times$11$\times$11$\times$24 \\\cline{2-3}
                          & ResBlock                    &11$\times$11$\times$11$\times$36 \\
\hline
\hline
                          & Concat with control param.      &11$\times$11$\times$11$\times$40 \\\cline{2-3}
                          & ResBlock                             &11$\times$11$\times$11$\times$40 \\
\hline
\hline
\multirow{6}{*}{Decoder}  & ResBlock                    &11$\times$11$\times$11$\times$36 \\\cline{2-3}
                          & ReLU(BN(ConvTrans(2x2x2, stride=2)))               &22$\times$22$\times$22$\times$24 \\\cline{2-3}
                          & ResBlock                    &22$\times$22$\times$22$\times$24 \\\cline{2-3}
                          & ReLU(BN(ConvTrans(2x2x2, stride=2)))               &44$\times$44$\times$44$\times$16 \\\cline{2-3}
                          & ResBlock                    &44$\times$44$\times$44$\times$16 \\\cline{2-3}
                          & ReLU(BN(ConvTrans(2x2x2, stride=2)))               &88$\times$88$\times$88$\times$8  \\
\hline
\hline
\multirow{3}{*}{\makecell{ Prediction}} 
                          & ResBlock                    & 88$\times$88$\times$88$\times$4  \\\cline{2-3}
                          & Dropout(ReLU(BN(Conv(1x1x1, 4$\rightarrow$4)))) & 88$\times$88$\times$88$\times$4  \\\cline{2-3}
                          & Conv(1x1x1, 4$\rightarrow$1)                    & 88$\times$88$\times$88$\times$1  \\

\hline 
\end{tabular}
\label{table:layers} 
\end{table}
\end{document}